\documentclass{article}

\usepackage{arxiv}

\usepackage[utf8]{inputenc} 
\usepackage[T1]{fontenc}    
\usepackage{hyperref}       
\usepackage{url}            
\usepackage{booktabs}       
\usepackage{amsfonts}       
\usepackage{nicefrac}       
\usepackage{microtype}      
\usepackage{lipsum}
\usepackage{bbm}
\usepackage{xcolor}
\usepackage{enumitem}
\usepackage{amsmath,epsfig}
\usepackage{amssymb}
\usepackage{amsthm}
\usepackage{algorithm}
\usepackage{algpseudocode}
\usepackage{tikz}
\usepackage{tabu}
\usepackage{graphicx}
\usepackage{subfig}
\usepackage{epstopdf}
\usepackage{epsfig}
\usepackage{booktabs}
\usepackage[flushleft]{threeparttable}
\usetikzlibrary{tikzmark,calc}
\usepackage{float}
\usepackage{multirow}
\usepackage{hyperref}
\usepackage{verbatim}
\usepackage{wrapfig}
\usepackage{rotating}
\usepackage{mathtools}
\usepackage{bm}
\usepackage{bbm}
\usepackage{enumitem}
\usepackage{standalone}
\usepackage{pgfplots}
\usepackage{pgfplotstable}
\numberwithin{equation}{section} 

\usepackage{microtype}

\def\R{{\mathbb{R}}}

\newcommand{\etal}{\textit{et al.}}

\newcommand{\Eqref}[1]{Eq.~\ref{#1}} 
\newcommand{\Algref}[1]{Alg.~\ref{#1}}
\newcommand{\Secref}[1]{Sec.~\ref{#1}}
\newcommand{\Figref}[1]{Fig.~\ref{#1}}

\pgfdeclarelayer{background}
\pgfdeclarelayer{foreground}
\pgfsetlayers{background,main,foreground}

\tikzstyle{sensor}=[draw, fill=blue!20, text width=4em, 
    text centered, minimum height=2.5em]
\tikzstyle{normalCell}=[draw, fill=orange!20, text width=3.2em, 
    text centered, minimum width=2.0em, minimum height=1.4em, 
    rounded corners]
\tikzstyle{conv}=[draw, fill=blue!20, text width=4.2em, 
    text centered, minimum width=2.0em, minimum height=1.4em, rounded corners]
\tikzstyle{operation}=[draw, fill=white!20, text width=3.2em, 
    text centered, minimum width=2.0em, minimum height=1.4em, rounded corners]
\tikzstyle{cell_in}=[draw, fill=green!20, text width=3.0em, 
    text centered, minimum width=2.5em, minimum height=2.3em]
\tikzstyle{cell_out}=[draw, fill=yellow!20, text width=3.0em, 
    text centered, minimum width=2.5em, minimum height=2.3em]
\tikzstyle{square}=[draw, fill=blue!20, minimum size=2em]
\tikzstyle{choice}=[draw, fill=green!20, text width=3.2em, 
    text centered, minimum width=2.0em, minimum height=1.4em, rounded corners]
\tikzstyle{customNode}=[draw, fill=purple!25, text width=3.2em, 
    text centered, minimum width=2.0em, minimum height=1.4em, rounded corners]
\tikzstyle{ann} = [above, text width=5em, text centered]
\tikzstyle{wa} = [sensor, text width=10em, fill=red!20, 
    minimum height=6em, rounded corners, drop shadow]
\tikzstyle{sc} = [sensor, text width=13em, fill=red!20, 
    minimum height=10em, rounded corners, drop shadow]
\tikzstyle{texto} = [above, text width=6em, text centered]
\tikzstyle{oneShotModel}=[draw, fill=white!10, text width=3.0em, 
    text centered, minimum width=2.0em, minimum height=6.0em]
\tikzstyle{sampleNode}=[draw, fill=white!10, text width=5.0em, 
    text centered, minimum width=2.0em, minimum height=1.0em]
\tikzstyle{CSNode}=[draw, fill=white!10, text width=4.0em, 
    text centered, minimum width=2.0em, minimum height=6.0em]
\tikzstyle{restrictionNode}=[draw, fill=white!10, text width=5.0em, 
    text centered, minimum width=2.0em, minimum height=1.0em]

 \newcommand{\backgroundc}[6]{%
  \begin{pgfonlayer}{background}
    \path (#1.west |- #2.north)+(-0.3, 1.28) node (a1) {};
    \path (#3.east |- #4.south)+(+0.3,-1.42) node (a2) {};
    \path[fill=#5!20,rounded corners, draw=black!50, dashed]
      (a1) rectangle (a2);
    \path (a1.west |- a1.north)+(1.1, -0.65) node (u1)[texto]
      {\small #6};
  \end{pgfonlayer}}
  
\newcommand{\backgroundd}[6]{%
  \begin{pgfonlayer}{background}
    \path (#1.west |- #2.north)+(-0.3, 0.5) node (a1) {};
    \path (#3.east |- #4.south)+(+0.3,-0.5) node (a2) {};
    \path[fill=#5!20,rounded corners, draw=black!50, dashed]
      (a1) rectangle (a2);
    \path (a1.west |- a1.north)+(2.3, -0.65) node (u1)[texto]
      {\small #6};
  \end{pgfonlayer}}

\title{ESPN: Extremely Sparse Pruned Networks}

\newcommand{\rvw}{\mathbf{w}}
\newcommand{\rvc}{\mathbf{c}}

\author{%
  Minsu Cho \qquad Ameya Joshi \qquad Chinmay Hegde\thanks{This work was supported in part by grants CCF-2005804 and CCF-1815101 from the National Science Foundation.}\\
  Tandon School of Engineering\\
  New York University \\
  \texttt{\{mc8065, aaj458, chinmay.h\}@nyu.edu} 
}

\begin{document}

\maketitle

\begin{abstract}
    Deep neural networks are often highly overparameterized, prohibiting their use in compute-limited systems. However, a line of recent works has shown that the size of deep networks can be considerably reduced by identifying a subset of neuron indicators (or mask) that correspond to significant weights prior to training. We demonstrate that an simple iterative mask discovery method can achieve state-of-the-art compression of very deep networks. Our algorithm represents a hybrid approach between single shot network pruning methods (such as SNIP) with Lottery-Ticket type approaches. We validate our approach on several datasets and outperform several existing pruning approaches in both test accuracy and compression ratio. 

\end{abstract}

\section{Introduction}
\label{sec:intro}

\noindent\textbf{Motivation.}  Neural networks have achieved state of the art results across several domains such as computer vision, language processing, and reinforcement learning. This performance is generally contingent on large, over-parameterized networks that are trained using massive amounts of data. For example, the current state of the art on the ImageNet classification task uses a network with over 480 million parameters~\cite{Touvron2020FixingTT}, and consequently, the best performing networks are prohibitive in terms of computational and memory requirements. Therefore, compressing neural networks is vital for resource limited settings (such as self-driving cars and mobile devices). 
In this work, we present an algorithm for compressing neural networks to far higher degree of sparsity levels than has been reported in the literature. 

\noindent\textbf{Challenges.} Network pruning involves finding a smaller, more efficient representation of a given reference neural network. Pruning strategies include  quantization~\cite{han2015deep, hubara2017quantized}, sparsification~\cite{NIPS1988_133,tartaglione2018learning, dai2018compressing, Frankle2018TheLT, Lee2019SNIPSN}, and distillation~\cite{belagiannis2018adversarial, sau2016deep, polino2018model}. See~\cite{cheng2017survey} for a detailed look at several approaches towards neural network compression.

In sparsification-based network pruning, the goal is to select a subset of the original weights that are easily encoded while preserving nominal performance on a test set. 
However, such methods involve expensive \emph{retraining} of the weights once the subset is identified, as well as requires significant hyper-parameter tuning. \cite{Lee2019SNIPSN} propose a method called {SNIP}, which identifies a salient subset of weights for a given dataset \emph{prior} to training. This avoids re-training of the weights and thereby offers computational advantage over existing methods, but achieves weaker compression performance as a trade-off. 

The authors of~\cite{Frankle2018TheLT} present the \emph{Lottery Ticket hypothesis}, wherein they demonstrate the existence of randomly initialized sub-networks whose weights can be optimized to achieve high performance. Subsequent work by \cite{Frankle2019TheLT} stabilizes the algorithm for finding a winning `ticket' (i.e., a good sub-network) via iterative pruning methods. However, iterative methods incur larger computation compare to single-shot methods such as \cite{Lee2019SNIPSN}. \cite{zhou2019deconstructing} further suggest that finding the winning ticket for very deep networks is an implicit consequence of the training mechanism.  In any case, all these methods require higher computational costs, as well as significant hyperparameter tuning overhead, than those incurred by a baseline training approach.

\noindent\textbf{Contributions.} Given these observations, clear trade-offs exist among computation, sensitivity to hyperparameters, and final test performance. In this paper, we show that it is possible to achieve excellent pruning results with not only low computational costs, but also while being robust to tuning hyperparameters. 
In essence, our method merges \cite{Lee2019SNIPSN}'s {single-shot pruning} approach with the {iterative pruning} strategy by \cite{zhou2019deconstructing}. We demonstrate that our approach is successful at pruning well-initialized, large networks such as residual networks (ResNets). Somewhat surprisingly, our empirical results demonstrate that it may be possible to compress neural networks to considerably higher pruning levels than has been reported so far in the literature.

Our specific contributions are fourfold: 
\begin{enumerate}[nosep,leftmargin=*]
    \item We present an algorithm for training extremely sparse neural networks, by learning masking operators for the weights using gradient updates through convex relaxation.
    \item We achieve a 5X improvement over baseline methods such as SNIP, while conserving accuracy.
    \item We provide extensive comparison on our approach on a variety of models and datasets, outperforming various state-of-the-art sparse pruning methods especially in extreme pruning ratio ($>99\%$).
    \item Finally, we conduct ablation studies and hyperparameter sensitivity experiments to analyse each component of our approach in detail. 
\end{enumerate}


\section{Related Work}
\label{sec:rel_work}


Early neural network compression approaches \cite{han2015deep,collins2014memory,guo2016dynamic} rely on simple magnitude-based heuristic pruning tactics and show significant compression of deep networks while preserving performance. \cite{han2015deep}, \cite{guo2016dynamic}, and \cite{zhu2017prune} rely on iterative fine-tuning and pruning to reduce the effect of removing weights on the output. Other work adopts second-order approximation of loss surface (\cite{lecun1990optimal}, \cite{hassibi1993second}) to minimize performance degradation, but do not scale well to larger models.
Leveraging techniques from information theory and Bayesian statistics have further improved model compression ratios, including information bottlenecks \cite{dai2018compressing}, Fisher pruning \cite{molchanov2016pruning, theis2018faster}, entropy-penalized reparameterization \cite{oktay2019model}, and Variational Bayes \cite{molchanov2017variational, louizos2017bayesian, ullrich2017soft}.  

While most of the above approaches are applied to pre-trained existing networks, there has been a recent surge in approaches that interleave pruning and training. Such approaches rely on reparameterizing the weights and training over the modified network. Methods such as low-rank decomposition~\cite{denil2013predicting}, FastFood transforms~\cite{Yang2015DeepFC}, hashing~\cite{chen2015hashednet} leverage reparameterization schemes for fully connected layers. An obvious reparameterization is inducing sparsity on model connections. However, training sparse models can be unstable. To resolve such issues, \cite{mocanu2018scalable} and \cite{dai2018compressing} use iterative parameter growth, allowing the sparse architecture additional degrees of freedom. Evolutionary approaches such as \cite{bellec2017deep} propose a dynamic sparsification model using connectivity constraints to train an implicitly sparse model. \cite{mostafa2019parameter} further improve this by allowing a global sparsity constraint instead of a layer-wise approach, thus forcing a network to learn a consistent sparse model.

Another branch of pruning algorithms learns the auxiliary (masking) parameters via gradient-based approaches by relaxing non-differentiable constraints (e.g.,  $\ell_0$-constraints) to differentiable estimators. \cite{Louizos2018LearningSN} re-parameterizes the weights with the element-wise product of its weight and auxiliary parameters sampled by learning the hard concrete distribution. \cite{xiao2019autoprune} adopts the straight-through estimator to take gradients with respect to the indicator function which generates the binary masks to train auxiliary parameters. \cite{zhou2019deconstructing} and \cite{savarese2019winning} trains the auxiliary parameters through gradient descent by reparameterizing with Bernoulli samplers with sigmoidal probabilities on the auxiliary parameters. \cite{sanh2020movement} adopts the hard-concrete distribution proposed in \cite{Louizos2018LearningSN} to learn both weights and auxiliary parameters simultaneously. We note that our work and the recent preprint \cite{sanh2020movement} share similar concepts of updating weights and masks simultaneously, but take different approaches for treating non-differentiable regularized terms.  



\cite{lin2017runtime} present a runtime pruning method that models the flow of activations through the network as a Markov process, and use reinforcement learning to learn the best policy for pruning. \cite{he2018amc} also use reinforcement learning in the context of AutoML to train model architectures that are explicitly optimized for edge devices.

Instead of pruning in the middle of or after the training, \cite{Lee2019SNIPSN, wang2020picking, verdenius2020pruning} proposes the method pruning network prior to the training based on saliency scores from an untrained network. While pruning an untrained network provides a huge advantage in computational cost, the pruning algorithms on a trained network provides par or better compressing performances than single-shot prunings on untrained networks.  


While pruning weights in the level of each weight (unstructured) has the advantage of compressing the network in the highest degree of freedom, the limitation on inference speed exist due to limited support in current deep learning software frameworks on parallel processing in each weight level. Instead, structured pruning approaches~\cite{liu2017learning, liu2018rethinking, luo2017thinet, gray2017gpu, anwar2017structured} focuses pruning in a higher structured level such as neurons or filters such that corresponding weight matrix only contains zero elements. In this work, we focus on unstructured weight pruning by targeting competitive sparsity-accuracy tradeoff. 

\section{Preliminaries}
\label{sec:snip_lth}

\noindent\textbf{Notation.} Let $f:\R^d\to\R^k$ be a neural network with weights $\rvw \in \R^m$. Assume $c_i\in\{0, 1\}$, an auxiliary indicator for each of the weight, where each $c_i$ indicates if the weight $w_i$ is a \emph{salient} parameter. Saliency here refers to the effect of zeroing out $w_i$ on the final risk. A high saliency value (greater than some threshold $\epsilon$) will be indicated by $c_i = 1$. 

To motivate our proposed approach, we first describe two existing pruning methods.

\noindent\textbf{SNIP.} Single-Shot Network Pruning (SNIP)~\cite{Lee2019SNIPSN} introduces a saliency based criterion to prune network connections. The premise is that since neural networks are often overparameterized, they  have redundant connections which can be identified prior to training using a saliency-based approach. 

Given a dataset $\mathcal{D} = \{\mathbf{x}, \mathbf{y}\}$, and a desired network sparsity level $k$, training a sparse neural network amounts to solving the following optimization problem:
\begin{eqnarray}
    \min_{\rvw} L(f(\rvw; \mathbf{x}),  \mathbf{y}), ~~ \text{s.t.}~~\| \rvw \|_0 \leq k.
\end{eqnarray}
Here, $L(\cdot)$ refers to any standard loss function, such as the cross entropy or $\ell_2$ loss. Enforcing the $\ell_0$-constraint is combinatorially difficult but can be relaxed either via a sparsifying penalty~\cite{carreira2018learning, NIPS1988_133, setiono1997penalty, Louizos2018LearningSN} or a saliency-based weight dropping mechanism~\cite{han2015deep, NIPS1988_119}.  

SNIP instead uses the following alternative formulation using an auxiliary indicator variable, $\rvc \in \{0, 1\}^m$ that indicates if a weight contributes (positively or negatively) to the output. The modified loss therefore is as follows, 
\begin{equation}
    \min_{\rvw, \rvc} L(f(\rvc \odot \rvw; \mathbf{x}), \mathbf{y}),~~\text{s.t.}~~ \|\rvc\|_0 \leq k. 
    \label{eq:snip_loss}
\end{equation}
Here, $\odot$ refers to element-wise multiplication. Observe that the auxiliary variable decouples the constraint from the weights, $\rvw$. This decoupling allows for measuring the saliency of a specific weight matrix by measuring the effect of removing a connection. For example, if we consider the difference in loss by removing the $j^{th}$ edge:
\begin{equation}
    \Delta L_j = L(\mathbf{1} \odot \rvw) - L((\mathbf{1} - \mathbf{e}_j) \odot \rvw)
\end{equation}
where $e_j$ is an indicator vector, then the absolute value of $\Delta L_j$ indicates the contribution of the $j^{th}$ parameter to the  performance of the network. 
Instead of directly computing $\Delta L$ for $m$ weights which involves cumbersome forward passes, the author approximates $\Delta L$ with the gradient, $\partial L/\partial c_j$, which can be easily calculated using automatic differentiation.
\emph{A posteriori}, only the top $k$ connections with the highest gradient values are retained. This can be done using a simple thresholding operation. Specifically, $c_j$ is defined as the following indicator variable (denoting connection sensitivity),
\begin{equation}
    c_j = \mathbbm{1}\left[s_j - s_k \geq 0\right],~~\text{where}~~s_j = \frac{\partial L/\partial c_j}{\sum_i \partial L / \partial c_i},
\end{equation}
and $s_k$ corresponds to the $k^{th}$ largest sensitivity value.  
The network weights are then randomly initialized and further trained with the $c_j$'s kept constant. This allows for a single-shot compression scheme with only one round of training. 

\noindent\textbf{Lottery Ticket Hypothesis.} The Lottery Ticket (LT) hypothesis proposed by \cite{Frankle2018TheLT} claims the existence of a smaller subnetwork within a standard large, dense neural network architecture that will provide competitive performance when trained from scratch as that of the original network. This suggests the need for network pruning to become an essential component of the training process. However, in order to find such a subnetwork, the authors propose an iterative process that alternately prunes and retrains the pruned network from scratch. They also introduce `rewinding', where one retrains the pruned model starting with the initial (random) weights instead of the final learned weights upon which the pruning is performed. 

\cite{zhou2019deconstructing} further systematically analyse the LT hypothesis and provide two important observations: (1) new values on kept weights should have the same sign as the original initial values (in line with why rewinding in \cite{Frankle2018TheLT} is important), (2) it is important to set to masked weights to zero, instead of using values other than zero. From the second observation, \cite{zhou2019deconstructing} suggests that magnitude-based masking criteria (which \cite{Frankle2018TheLT} use) tends to prune weights that seem to move towards zero during training. 



\begin{algorithm}[t!]
    \small
    \caption{\textsc{ESPN-Finetune}}
    \begin{algorithmic}[1]
        \State\textbf{Inputs: } $f(\rvw)$: pre-trained network ($\rvw \in \mathbb{R}^d$), $\rvc=\mathbf{1}$: auxiliary parameters, $T$: fine-tuning epochs, $\alpha$: sparsity weight, $\eta$: learning rate, $p$: pruning ratio, $\epsilon$: threshold. 
        \Procedure{Training $w$ and $c$ via SGD}{}
            \State $\overline{\rvw} \gets (\rvw, \rvc)$
            \While {$N_c \geq d \cdot (1-p)$}
                \State $\overline{\rvw} \gets \overline{\rvw} - \eta \nabla_{\overline{\rvw}} (L(f(\rvw \odot \rvc; x), y) + \alpha \|\rvc\|_1)$
                \State $N_c \gets$ SUM($\mathbbm{1}(\rvc > \epsilon)$)
            \EndWhile
        \State $\rvw\gets \rvw \odot \rvc$
        \State $\rvc\gets \mathbbm{1}(\rvc > \epsilon)$
        \EndProcedure
        \Procedure{Fine-tune the Network}{}
            \State {Train $f(\rvw \odot \rvc)$ respect to w for $T$ epochs.}
        \EndProcedure
    \end{algorithmic}
    \label{alg:finetune}
\end{algorithm}

\begin{algorithm}[t!]
    \caption{\textsc{ESPN-Rewind}}
    \begin{algorithmic}[1]
        \State\textbf{Inputs: } $f(\rvw)$: Untrained Network ($\rvw \in \mathbb{R}^d$), $t$: warmup epochs, $T$: epochs $\rvc=\mathbf{1}$: auxiliary parameters, $\alpha$: Lasso coefficient, $\eta$: learning rate, $p$: pruning ratio, $\epsilon$: threshold. 
        \Procedure{Warmup Training}{}
            \For {epoch $\in \{1,\ldots, t\}$}
                \State $\rvw \gets =\rvw - \eta \nabla_{\rvw}(L(f(w)))$
            \EndFor
            \State $\rvw_t \gets \rvw$
        \EndProcedure
        \Procedure{Training $\rvw$ and $\rvc$ via SGD}{}
            \State $\overline{\rvw} \gets (\rvw, \rvc)$
            \While {$N_c \geq \text{} d \cdot (1-p)$}
                \State $\overline{\rvw} \gets \overline{\rvw} - \eta \nabla_{\overline{\rvw}} (L(f(\rvw \odot \rvc; x), y) + \alpha \|\rvc\|_1)$
                \State $N_c \gets$ SUM($\mathbbm{1}(\rvc > \epsilon)$)
            \EndWhile
        
        \State $\rvc\gets \mathbbm{1}(\rvc > \epsilon)$
        \EndProcedure
        \Procedure{Rewind and train the network}{}
            \State $\rvw \gets \rvw_t \odot \rvc$
            \State {Train $f(\rvw)$ respect to $\rvw$ via SGD for $T-t$ epochs}
        \EndProcedure
    \end{algorithmic}
    \label{alg:rewind}
\end{algorithm}

\section{Our Approach: ESPN}
\label{sec:espn}

While SNIP does enable very good network compression at moderate computational cost, a single-shot approach before training has several disadvantages. The primary issue is that connection sensitivities estimated using single-shot techniques may be erroneous. This may lead to discarding of network edges that eventually would lead to better network performance. 

To address these, we present a novel network pruning method: \emph{Extremely Sparse Pruned Networks} (ESPN). Our method resembles SNIP, but learns the sparse masking operator, $\rvc$, via a standard iterative gradient update framework, instead of using a single-shot estimator. This modification to the standard SNIP framework is also inspired by the observations from Zhou~\etal~\cite{zhou2019deconstructing} that learning such sparse indicators can be viewed as a natural process concurrent to training a neural network on data. 

\noindent\textbf{Approach.} Our approach consists of three steps: 
\begin{enumerate}[nosep,leftmargin=*]
    \item pretraining $\rvw$ while freezing $\rvc=\mathbf{1}$,
    \item leveraging a relaxed form of the SNIP saliency objective to train $\rvc$, thereby pruning the network to required sparsity, and finally,
    \item finetuning the pruned network to boost final performance.
\end{enumerate}

For the first step, we train our base architecture on the given dataset to ensure a good initialization for the sparse problem. However, we note that unlike previous works~\cite{han2015deep, guo2016dynamic, lecun1990optimal, hassibi1993second, Louizos2018LearningSN, molchanov2017variational}, which rely on a fully trained network as input, we do not require our network to be trained to convergence. Instead, we require the network to only be trained for a few iterations. We also point out that in case of preexisting networks, this step can be safely skipped.

We further modify the SNIP objective in order to iteratively learn masks. First, instead of freezing weights and indicators, we simultaneously train them both, thus keeping track of changes in sensitivity values during training. To achieve this, we modify any given architecture to have an additional matrix associated with each weight matrix such that $\mathbf{C} = \{\mathbf{C}^{i} = \mathbf{1}^{m \times n} | \mathbf{W}^{i} \in \mathbb{R}^{m \times n}\}$. This is similar to the implementation of the auxiliary variable in SNIP. Secondly, we relax the sparsity constraint in \Eqref{eq:snip_loss} to an $\ell_1$ penalty, so as to be make the objective differentiable. The training objective is given by:
\begin{equation}
    \min_{\rvw, \rvc} L(f(\rvc \odot \rvw, \mathbf{x}), \mathbf{y}) + \alpha \|\rvc\|_1 \, .
\label{eq:espn_loss}
\end{equation}
We rely on standard backpropagation (e.g. SGD) to update both $\rvw$ and $\rvc$, until we achieve the required sparsity. We propose a simple update for $\rvc$ with $\rvc \in \mathbb{R}^m$ initialized to $\mathbf{1}$ rather than randomly as in \cite{belagiannis2018adversarial, Louizos2018LearningSN, xiao2019autoprune, savarese2019winning, sanh2020movement, zhou2019deconstructing}. This is advised by Zhou~\etal~'s~\cite{zhou2019deconstructing} observations regarding the masking operation. Note that optimizing \Eqref{eq:espn_loss} with respect to $\rvc$, $\rvc$ may no longer be sparse with $c_i \notin \{0, 1\}$. Assuming that $\rvc$ is the optimal selection with salient connections, we update $\rvw$ via the element-wise product of $\rvw$ and $\rvc$ as per \Eqref{eq:espn_loss}. Subsequently, we restore $\rvc$ to be an indicator function by thresholding,  $\mathbbm{1}(\rvc > \epsilon)$ where $\epsilon$ is a hyperparameter corresponding to non-zero elements in $\rvc$.


For our algorithm, the choice of the termination condition is significant. Given a target pruning ratio $p$, we train both weights $\rvw$ and $\rvc$ until sparsity of $\rvc$ is less or equal to target sparsity $1-p$. While the alternative approach is to train $\rvw$ and $\rvc$ with a given fixed training budget, the algorithm may either not reach the required sparsity level or may unnecessarily waste computation. Our terminating condition allows us to not only to terminate the training, but also removes a sensitive hyperparameter (no. of epochs). 

For the third (finetuning) step, we consider two variants. The first variant simply trains the pruned network with the given dataset with a low learning rate until we achieve the desired accuracy (we call this \textsc{ESPN-Finetune}; see \Algref{alg:finetune}). 

Alternatively, we can also use the `rewinding' technique from \cite{Frankle2018TheLT} and \cite{renda2020comparing}. Rewinding involves training the pruned architecture by initializing weights from a previously well-performing supernetwork. In this case, we use the subset of $\rvw_t$ from the warm-up training (trained $t$ epochs) instead of pretrained weights. After learning auxiliary parameter $\rvc$ with a procedure from ~\Algref{alg:finetune} Line (4-9), we rewind to epoch $t$ updating weights by $\rvw= \rvw_t \odot \rvc$ and train the model with remaining budget.
We call this \textsc{ESPN-Rewind}; see \Algref{alg:rewind}.

In the following section, we present a comprehensive evaluation of our approach on various image classification networks and datasets. Our code can be accessed at \url{https://github.com/chomd90/extreme_sparse}.

\section{Experiments and Results}

We analyse the performance of our approach, ESPN, on various image classification tasks. For purposes of our analysis, we consider three architectures, (1) simple fully-connected networks, (2) VGG architectures, and (3) residual networks, for various pruning ratios. Specifically, we first consider fully connected networks trained for MNIST in order to demonstrate the efficacy of our approach. Subsequently, we study the efficacy of ESPN for compressing two massively overparameterized architectures, VGG and ResNet trained on complex datasets such as CIFAR-10/100~\cite{cifar10} and ImageNet~\cite{ILSVRC15}. 

We compare our approach with several pruning methods: (1) SNIP~\cite{Lee2019SNIPSN}, GraSP~\cite{wang2020picking}, (2) stabilized Lottery Ticket Hypothesis (LT)~\cite{Frankle2019TheLT} and (3) Dynamic Sparse Reparameterization (DSR)~\cite{mostafa2019parameter}. Additionally, we conduct ablation studies to understand the roles of various algorithm components. 

\noindent\textbf{Experimental Setup.} To ensure fair comparisons, we run all algorithms with official implementations and with the best hyperparameters reported in the literature. In the case of \textsc{ESPN-Rewind}, for all experiments except ImageNet, we train the network for 160 epochs through SGD with learning rate 0.1, momentum parameter 0.9, and weight decay 0.0005. We also decay the learning rate with a factor of 0.1 at epochs 80 and 120. For ImageNet, we adapt the official PyTorch implementation \cite{pytorch} to train the pruned network for \textsc{ESPN-Rewind} without modifying the hyperparameter setups (90 epochs, learning rate 0.1, learning rate decay 0.1 every 30 epochs, and weight decay 0.0001).

For \textsc{ESPN-Finetune}, we subsequently train the network for 50 epochs with SGD with a learning rate of 0.001 with a decay factor of 0.1 at epoch 30. We also use the weight decay coefficient with 0.0005 for all training except ImageNet. We use the pretrained ResNet50 from Pytorch library. For ESPN finetuning stage, we train ResNet50 on ImageNet with 2/3 of training epochs to the official pytorch implementation (60 epoch, learning rate 0.01, learning rate decay 0.1 at 30 and 50 epoch). For LT, we prune the fully-trained network with respect to magnitude and rewind to the early epoch as suggested in \cite{renda2020comparing}. 

While we train both model weights and auxiliary parameters, we use a standard SGD with Nesterov-momentum 0.9, and no weight decay penalty.

\subsection{Experimental Results}
\label{subsec:results}

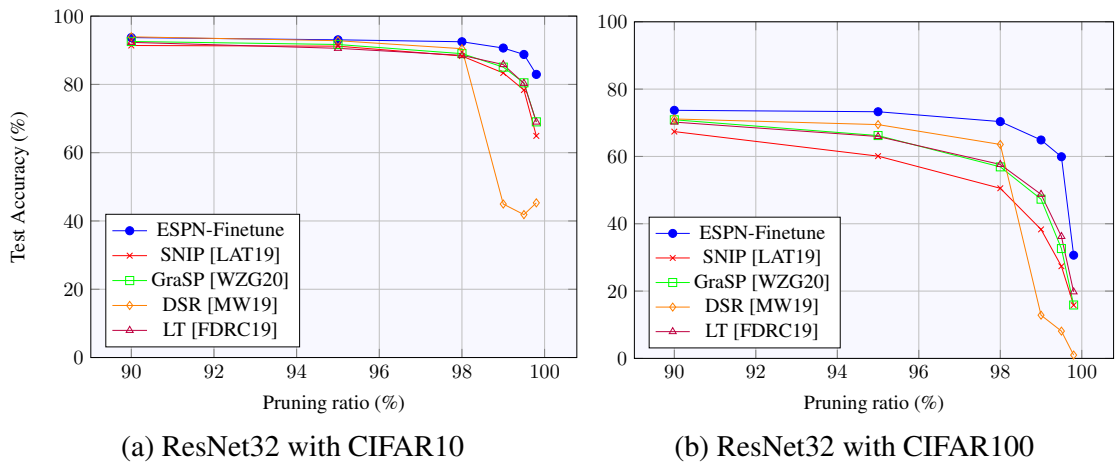
\begin{figure}[h!]
\centering
    \resizebox{.95\linewidth}{!}{
        \def\whist{0.37\linewidth}
        \def\whista{0.3380\linewidth}
        \begin{tabular}{c c}
        \subfloat[ResNet32 with CIFAR10]{\resizebox{\whist}{!}{\begin{tikzpicture}
    \begin{axis}[legend pos=south west,
        xlabel=Pruning ratio (\%),
        ylabel=Test Accuracy (\%),
        ymin=0, ymax=100,
        axis background/.style={fill=blue!3},
        grid=both,
        grid style={line width=.1pt, draw=gray!10},
        major grid style={line width=.2pt,draw=gray!50},
        width=10cm,height=7.5cm]
    \addplot[mark=*,blue] 
        plot coordinates {
            (90,93.71)
            (95,93.06)
            (98,92.49)
            (99,90.65)
            (99.5,88.77)
            (99.8,82.93)
        };
    \addlegendentry{ESPN-Finetune}

    \addplot[mark=x,red] 
        plot coordinates {
            (90,91.39)
            (95,91.20)
            (98,88.31)
            (99,83.35)
            (99.5,78.36)
            (99.8,64.92)
        };
    \addlegendentry{SNIP~\cite{Lee2019SNIPSN}}

    \addplot[mark=square,green] 
        plot coordinates {
            (90,92.57)
            (95,91.69)
            (98,89.01)
            (99,85.06)
            (99.5,80.46)
            (99.8,69.03)
        };
    \addlegendentry{GraSP~\cite{wang2020picking}}

    \addplot[mark=diamond,orange] 
        plot coordinates {
            (90,93.93)
            (95,92.80)
            (98,90.46)
            (99,44.96)
            (99.5,41.86)
            (99.8,45.31)
        };
    \addlegendentry{DSR~\cite{mostafa2019parameter}}
    
    \addplot[mark=triangle,purple] 
        plot coordinates {
            (90,92.31)
            (95,90.57)
            (98, 88.51)
            (99,85.81)
            (99.5,80.31)
            (99.8,68.96)
        };
    \addlegendentry{LT~\cite{Frankle2019TheLT}}
        
    \end{axis}
\end{tikzpicture}}}
        \vspace{0.7em}
        \subfloat[ResNet32 with CIFAR100]{\resizebox{\whista}{!}{\begin{tikzpicture}
    \begin{axis}[legend pos=south west,
        xlabel=Pruning ratio (\%),
        ymin=0, ymax=100,
        axis background/.style={fill=blue!3},
        grid=both,
        grid style={line width=.1pt, draw=gray!10},
        major grid style={line width=.2pt,draw=gray!50},
        width=10cm,height=7.5cm]
    \addplot[mark=*,blue] 
        plot coordinates {
            (90,73.70)
            (95,73.28)
            (98,70.35)
            (99,64.89)
            (99.5,59.91)
            (99.8,30.66)
        };
    \addlegendentry{ESPN-Finetune}

    \addplot[mark=x,red] 
        plot coordinates {
            (90,67.38)
            (95,60.09)
            (98,50.54)
            (99,38.32)
            (99.5,27.38)
            (99.8,15.82)
        };
    \addlegendentry{SNIP~\cite{Lee2019SNIPSN}}

    \addplot[mark=square,green] 
        plot coordinates {
            (90,70.89)
            (95,66.20)
            (98,56.90)
            (99,47.30)
            (99.5,32.63)
            (99.8,15.87)
        };
    \addlegendentry{GraSP~\cite{wang2020picking}}

    \addplot[mark=diamond,orange] 
        plot coordinates {
            (90,71.13)
            (95,69.46)
            (98,63.56)
            (99,12.84)
            (99.5,8.11)
            (99.8,1.00)
        };
    \addlegendentry{DSR~\cite{mostafa2019parameter}}
    
    \addplot[mark=triangle,purple] 
        plot coordinates {
            (90,70.20)
            (95,65.92)
            (98, 57.62)
            (99,48.74)
            (99.5,36.22)
            (99.8,19.77)
        };
    \addlegendentry{LT~\cite{Frankle2019TheLT}}
        
    \end{axis}
\end{tikzpicture}}}
        \end{tabular}}
    \caption{\sl \small Sparsity-accuracy tradeoff curve on ResNet32 with CIFAR10/100 dataset. Tested on extreme pruning ratios: \{90\%, 95\%, 98\%, 99\%, 99.5\%, 99.8\%\}. $99.8\%$ pruned ResNet32 has only 3.8k non-zero parameters. Note that ESPN-finetuning outperforms all other methods for extremely high sparsity levels while being comparable at lower sparsity levels.} 
    \label{fig: Pareto Curve}
\end{figure}

\begin{table*}[t!]
\centering
\small
\caption{\sl \textbf{LeNet300 and LeNet5-Caffe Comparison on MNIST and Fashion-MNIST}. Note that ESPN outperforms all other approaches for various pruning ratios except on LeNet5-Caffe trained withFashion-MNIST dataset pruning 95\% ratio.}
\begin{threeparttable}
\begin{tabular}{c c c c c c c c c c}
    \toprule
    \multicolumn{1}{l}{\textbf{Dataset}} & \multicolumn{4}{c}{MNIST} & \multicolumn{4}{c}{Fashion-MNIST} \\
    \midrule
    \multicolumn{1}{l}{\textbf{LeNet300-100}} & \multicolumn{2}{c}{Acc: 98.80\%}  &  \multicolumn{2}{c}{Params: 266K}  & \multicolumn{2}{c}{Acc: 89.81\%} & \multicolumn{2}{c}{Params: 266K} \\ 
    \midrule
    \multicolumn{1}{l}{\textbf{Pruning Ratio}} & 95\% & 98\% & 99\% & 99.6\% & 95\%& 98\% & 99\%& 99.6\% \\
    \multicolumn{1}{l}{} & (13K) & (5.3K) & (2.7K) & (1.1K) & (13K) & (5.3K) & (2.7K) & (1.1K) \\
    \midrule
    \multicolumn{1}{l}{SNIP} & 97.86 & 97.15 & 95.27 & 86.57 & 88.31 & 87.14 & 81.93 & 68.60 \\ 
    \multicolumn{1}{l}{GraSP} & 97.96 & 96.96 & 93.54 & 45.02 & 88.47 & 86.59 & 77.62 & 38.37  \\ 
    \multicolumn{1}{l}{LT\textsuperscript{+} (Rewind)} & 98.26 & 97.74 & 96.95 & 92.90 & 89.55 & 88.59 & 87.38 & 83.57 \\
    \midrule
    \multicolumn{1}{l}{ESPN-Rewind} & \textbf{98.57} & \textbf{98.39} & 97.94 & 97.24 & \textbf{89.94} & \textbf{89.33} & \textbf{88.87} & \textbf{87.74} \\ 
    \multicolumn{1}{l}{ESPN-Finetune} & 98.52 & 98.35 & \textbf{98.24} & \textbf{97.28} & 89.59 & 88.53 & 88.16 & 87.67\\ 
    \midrule
    \midrule
    \multicolumn{1}{l}{\textbf{LeNet5-Caffe}} & \multicolumn{2}{c}{Acc: 99.32\%}  &  \multicolumn{2}{c}{Params: 431K}  & \multicolumn{2}{c}{Acc: 90.48\%} & \multicolumn{2}{c}{Params: 431K} \\ 
    \midrule
    \multicolumn{1}{l}{\textbf{Pruning Ratio}} & 95\% & 98\% & 99\% & 99.6\% & 95\%& 98\% & 99\%& 99.6\% \\
    \multicolumn{1}{l}{} & (22K) & (8.6K) & (4.3K) & (1.7K) & (22K) & (8.6K) & (4.3K) & (1.7K) \\
    \midrule
    \multicolumn{1}{l}{SNIP} & 99.33 & 99.06 & 98.93 & 97.24 & 90.89 & 90.30 & 89.69 & 84.35 \\ 
    \multicolumn{1}{l}{GraSP} & 99.26 & 99.21 & 98.47 & 97.08 & 90.58 & 90.43 & 89.25 & 85.79\\ 
    \multicolumn{1}{l}{LT\textsuperscript{+} (Rewind)} & 99.27 & \textbf{99.27} & 99.17 & 98.30 & \textbf{91.57} & 91.11 & 90.36 & 87.96 \\
    \midrule
    \multicolumn{1}{l}{ESPN-Rewind} & \textbf{99.37} & 99.26 & \textbf{99.22} & 99.06 & 91.48 & 91.15 & 91.20 & 90.37 \\ 
    \multicolumn{1}{l}{ESPN-Finetune} & 99.30 & 99.26 & 99.10 & \textbf{99.10} & 91.22 & \textbf{91.48} & \textbf{91.60} & \textbf{90.94} \\ 
    \bottomrule
\end{tabular}
\begin{tablenotes}
        \small
        \item \textsuperscript{+} LT rewinded to epoch 1 after magnitude-based pruning on fully-trained networks.
    \end{tablenotes}
\end{threeparttable}
\label{table: lenet}
\end{table*}

\noindent\textbf{MNIST and Fashion-MNIST Dataset.} We start by evaluating ESPN for LeNet300 and LeNet5-Caffe. For the purposes of comparison, we use SNIP and GraSP as baselines.
LeNet300 and LeNet5-Caffe consist of 266K and 431K parameters respectively. 
We observe that the performance of ESPN to 95\% pruning ratio is at par or better when compared to other approaches as shown in Table~\ref{table: lenet}. Upon more severe pruning (99\% and 99.6\%), ESPN outperforms all SNIP, GraSP, and LT with only minor degradation in test accuracy. Surprisingly, ESPN-Finetune finds the ``lottery ticket'' on LeNet5-Caffe with Fashion-MNIST at $\mathbf{99.6\%}$ pruning ratio while achieving $\mathbf{90.94}\%$ test accuracy.


\begin{table*}[t!]
\centering
\caption{\sl \textbf{VGG19 and ResNet32: Comparison on CIFAR10 and CIFAR100}. ESPN compares favorably with DSR for the case of VGG-19 and CIFAR-10 while being better for the harder problem of CIFAR-100. For the more complicated ResNet32 model, ESPN outperforms other methods across all settings. DSR also fails to converge for extremely sparse settings ($>99\%$) in the case of ResNet model whereas our method significantly outperforms every other approach.}
\begin{threeparttable}
\begin{tabular}{c c c c c c c c c c c}
    \toprule
    \multicolumn{1}{l}{\textbf{Dataset}} & \multicolumn{4}{c}{CIFAR10} & \multicolumn{4}{c}{CIFAR100} \\
    \midrule
    \multicolumn{1}{l}{\textbf{VGG19}} & \multicolumn{2}{c}{Acc: 93.53\%}  &  \multicolumn{2}{c}{Params: 20M}  & \multicolumn{2}{c}{Acc: 73.96\%} & \multicolumn{2}{c}{Params: 20M} \\ 
    \midrule
    \multicolumn{1}{l}{\textbf{Pruning Ratio}} & 95\% & 98\% & 99\% & 99.5\% & 95\%& 98\% & 99\%& 99.5\% \\
    \multicolumn{1}{l}{} & (1M) & (400K) & (200K) & (100K) & (1M) & (400K) & (200K) & (100K) \\
    \midrule
    \multicolumn{1}{l}{SNIP~\cite{Lee2019SNIPSN}} & 92.97 & 92.37 & 10.00\textsuperscript{\#} & 10.00\textsuperscript{\#} & 71.90 & 19.60 & 1.00\textsuperscript{\#} & 1.00\textsuperscript{\#} \\ 
    \multicolumn{1}{l}{GraSP\cite{wang2020picking}} & 92.81 & 91.94 & 91.27 & 88.62 & 71.28 & 68.72 & 65.84 & 60.28  \\
    \multicolumn{1}{l}{DSR\cite{mostafa2019parameter}} & \textbf{94.00} & \textbf{93.57} & \textbf{93.15} & 91.62 & \textbf{72.96} & 70.77 & 69.70 & 66.79  \\
    \multicolumn{1}{l}{LT\textsuperscript{+}~\cite{Frankle2019TheLT}} & 93.15 & 92.70 & 91.29 & 10.00\textsuperscript{\#} & 70.97 & 69.13 & 66.32 & 17.60\\
    \midrule
    \multicolumn{1}{l}{ESPN-Rewind} & 93.57 & 92.72 & 91.88 & \textbf{91.94} & 71.68 & \textbf{70.85} & 69.48 & \textbf{67.93} \\ 
    \multicolumn{1}{l}{ESPN-Finetune} & 93.62 & 93.24 & 92.87 & 91.88 & 72.32 & 71.00 & \textbf{70.35} & 66.45 \\ 
    \midrule
    \midrule
    \multicolumn{1}{l}{\textbf{ResNet32}} & \multicolumn{2}{c}{Acc: 93.93\%}  &  \multicolumn{2}{c}{Params: 1.9M}  & \multicolumn{2}{c}{Acc: 74.83\%} & \multicolumn{2}{c}{Params: 1.9M} \\ 
    \midrule
    \multicolumn{1}{l}{\textbf{Pruning Ratio} } & 95\% & 98\% & 99\% & 99.5\% & 95\%& 98\% & 99\%& 99.5\% \\
    \multicolumn{1}{l}{} & (95K) & (38K) & (19K) & (9.5K) & (95K) & (38K) & (19K) & (9.5K) \\
    \midrule
    \multicolumn{1}{l}{SNIP~\cite{Lee2019SNIPSN}} & 91.20 & 88.31 & 83.35 & 78.36 & 63.82& 54.09 & 38.32 & 27.38   \\ 
    \multicolumn{1}{l}{GraSP~\cite{wang2020picking}} & 91.69 & 89.01 & 85.06 & 80.46 & 66.20 & 56.90 & 47.30 & 32.63  \\ 
    \multicolumn{1}{l}{DSR~\cite{mostafa2019parameter}} & 92.80 & 90.46 & 44.96 & 41.86 & 69.46 & 63.56 & 12.84$^\star$ & 8.11$^\star$ & \\ 
    \multicolumn{1}{l}{LT\textsuperscript{+}~\cite{Frankle2019TheLT}} & 90.57 & 88.51 & 85.81 & 80.31 & 65.92 & 57.62 & 48.74 & 36.22\\
    \midrule
    \multicolumn{1}{l}{ESPN-Rewind} & 91.83 & 90.54 & 89.93 & \textbf{89.31} & 70.76 & 69.42 & 64.83 & 56.88 \\ 
    \multicolumn{1}{l}{ESPN-Finetune} & \textbf{93.06} & \textbf{92.49} & \textbf{90.65} & 88.77 & \textbf{73.28} & \textbf{70.35} & \textbf{64.89} & \textbf{59.91} \\ 
    \bottomrule
\end{tabular}
    \begin{tablenotes}
        \small
        \item \textsuperscript{\#} Failed to converge with accuracy 1/(number of classes).
        \item \textsuperscript{$\star$} DSR failed to converge.
        \item \textsuperscript{+} LT rewinded to a epoch 5 after magnitude-based pruning on fully-trained networks.
    \end{tablenotes}
\end{threeparttable}
\label{table: cifar10/100}
\end{table*}

\noindent\textbf{CIFAR10/100 and Tiny-ImageNet/ImageNet Dataset.} We now evaluate ESPN on modern architectures, VGG19 and ResNet32 on CIFAR10/100, and Tiny-ImageNet image classification datasets. Note that the total number of parameters of VGG19 and ResNet32 are 20M and 1.9M respectively.\footnote{We use the ResNet architecture defined in \cite{wang2020picking} for our analysis.} 
For VGG19 with CIFAR10, we show comparable performance with DSR (the current state-of-the-art) for lower compression ratios. However, we outperform all other existing algorithms. Specifically, we draw attention to the high pruning ratio of $99.5\%$, where we report minimal degradation of accuracy for a highly sparse network with only $\sim100k$ parameters. We observe that ESPN outperforms SNIP, GraSP, LT, and DSR for all other cases except VGG19 with CIFAR10 shown in Table~\ref{table: cifar10/100}. Especially, all our candidate algorithms except ESPN faces huge degradation in performance when pruning extreme pruning ratio (99\% and 99.5\%). We notice that ESPN-Finetune represents the Pareto frontiers for ResNet32 on high pruning ratios: \{90\%, 95\%, 98\%, 99\%, 99.5\%, 99.8\%\} as shown in Figure~\ref{fig: Pareto Curve}. Overall, we observe that ESPN learns meaningful subnetworks with empirical evidence of achieving extreme compression ratios with minor accuracy degradation.

We observe similar performance on Tiny-ImageNet as that on CIFAR10/100 as seen in Table~\ref{table: tiny-imagenet}. While DSR achieved significantly higher test accuracy on 90\%-pruned VGG19 with CIFAR10, ESPN performs on par with DSR at higher pruning levels.
Similar to the CIFAR10/100 case, ESPN outperforms all other candidate algorithms for ResNet32 for three different pruning ratios. We note that for the highest pruning ratio (98\%), ESPN outperforms other approaches by a large margin.  

We further test our algorithm on ResNet50 (25.6M parameters) for the ImageNet dataset. We test two different pruning ratios \{80\%, 90\%\} using our approach and compare with reported results for SNIP, GraSP, and DSR. Note that our approach surpasses SNIP and GraSP for all pruning ratios, while being comparable to DSR. Specifically, \textsc{ESPN-Finetune} outperforms DSR in \emph{top-1} accuracy for the 80\% case, while being comparable in all other cases. 

\begin{table*}[t!]
\centering
\caption{\sl \textbf{VGG19 and ResNet32: Comparison on Tiny-Imagenet}. ESPN favorably surpasses state of the art approaches on most settings. Note that ESPN and DSR, being dynamic reparameterization methods are proven to exceed single-shot performance. ESPN also exceeds the performance of LT by a significant margin.}
\begin{threeparttable}
\begin{tabular}{c c c c | c c c c}
    \toprule
    \multicolumn{1}{l}{\textbf{Architecture}} & \multicolumn{3}{c}{VGG19: 61.70\% (20M)} & \multicolumn{3}{c}{ResNet32: 61.60\% (1.9M)} \\
    \midrule
    \multicolumn{1}{l}{\textbf{Pruning Ratio}} & 90\% & 95\% & 98\% & 90\% & 95\% & 98\% \\
    \multicolumn{1}{l}{} & (2M) & (1M) & (400K) & (190K) & (95K) & (38K) \\
    \midrule
    \multicolumn{1}{l}{SNIP~\cite{Lee2019SNIPSN}} & 61.07 & 57.12 & 0.5 & 51.56 & 40.41 & 24.81\\ 
    \multicolumn{1}{l}{GraSP~\cite{wang2020picking}} & 60.26 & 59.53 & 56.54 & 54.84 &48.45 & 37.25 \\ 
    \multicolumn{1}{l}{DSR~\cite{mostafa2019parameter}} & $\mathbf{62.43}^*$ & $59.81^*$ & 58.36\textsuperscript{*} & 57.19\textsuperscript{*} & 56.08\textsuperscript{*} & 12.42\textsuperscript{${\star}$} \\
    \multicolumn{1}{l}{LT\textsuperscript{+}~\cite{Frankle2019TheLT}} & 60.69 & 59.28 & 56.59 & 55.64 & 50.13 & 39.90\\
    \midrule
    \multicolumn{1}{l}{ESPN-Rewind} & 60.05 & \textbf{59.86} & \textbf{58.66} & 58.48 & 57.60 & 54.21  \\ 
    \multicolumn{1}{l}{ESPN-Finetune} & 59.46 & 59.27 & 57.67 & \textbf{60.08} & \textbf{59.83} & \textbf{54.56} \\ 
    \bottomrule
\end{tabular}
    \begin{tablenotes}
        \small
        \item \textsuperscript{*} Experimental results from \cite{wang2020picking}.
        \item \textsuperscript{$\star$} DSR failed to converge.
        \item \textsuperscript{+} LT rewinded to Epoch 5 after magnitude-based pruning on fully-trained networks.
    \end{tablenotes}
\end{threeparttable}
\label{table: tiny-imagenet}
\end{table*}

\begin{table}[t!]
\centering
\caption{\sl \textbf{ResNet50 on Imagenet}}
\begin{threeparttable}
\begin{tabular}{c c c | c c }
    \toprule
    \multicolumn{1}{l}{\textbf{Architecture}} & \multicolumn{4}{c}{ResNet50 (25.6m)} \\
    \midrule
    \multicolumn{1}{l}{\textbf{Pruning Ratio ($\kappa$)}} & \multicolumn{2}{c}{80\% (7.3m)} & \multicolumn{2}{c}{90\% (5.1m)} \\
    \midrule
    \multicolumn{1}{l}{\textbf{Test Accuracy}} & Top-1 & Top-5 & Top-1 & Top-5 \\
    \midrule
    \multicolumn{1}{l}{Unpruned Model} & 76.15 & 92.87 & - & - \\
    \multicolumn{1}{l}{SNIP\textsuperscript{*}} & 69.67 & 89.24 & 61.97 & 82.90 \\ 
    \multicolumn{1}{l}{GraSP\textsuperscript{*}} & 72.06 & 90.82 & 68.14 & 88.67 \\ 
    \multicolumn{1}{l}{DSR\textsuperscript{\#}} & 73.3 & 92.4 & 71.6 & 90.5 \\
    \midrule
    \multicolumn{1}{l}{ESPN-Rewind} & 72.60 & 91.08 & 68.70 & 89.00 \\ 
    \multicolumn{1}{l}{ESPN-Finetune} & \textbf{74.34} & 92.10 & 71.35 & \textbf{90.68} \\ 
    \bottomrule
\end{tabular}
    \begin{tablenotes}
        \small
        \item \textsuperscript{*} Experimental results from \cite{wang2020picking}.
        \item \textsuperscript{\#} Experimental results from \cite{mostafa2019parameter}.
    \end{tablenotes}
\end{threeparttable}
\label{table: imagenet}
\end{table}

\subsection{Visualizing Weight Distribution of Pruned Networks}
\label{subsection: weight distribution}

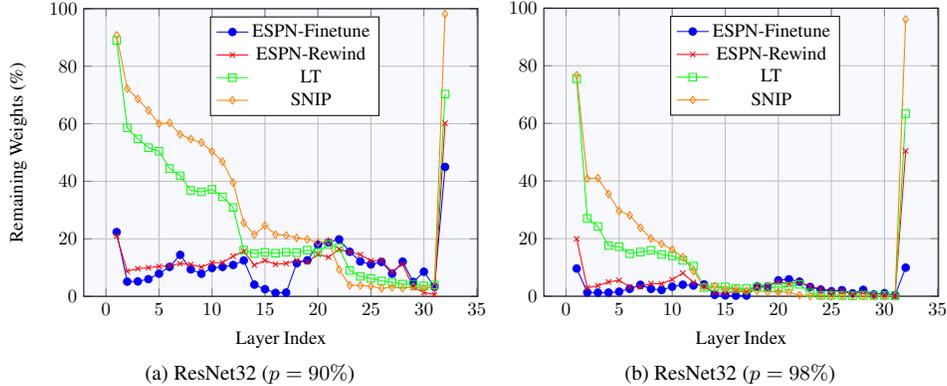
\begin{figure}[htp!]
\centering
    \resizebox{.8\linewidth}{!}{
        \begin{tabular}{c c}
        \vspace{-0.5em}
        \subfloat[ResNet32 ($p=90\%$)]{\resizebox{0.45\columnwidth}{!}{\begin{tikzpicture}
    \begin{axis}[ymin=0,ymax=100,
        legend style={at={(0.75,0.8)},anchor=east},
        xlabel=Layer Index,
        ylabel=Remaining Weights (\%),
        axis background/.style={fill=blue!3},
        grid=both,
        grid style={line width=.1pt, draw=gray!10},
        major grid style={line width=.2pt,draw=gray!50},
        width=9cm,height=7cm
        ]
    \addplot[mark=*,blue] 
        table {tikz/finetune_resnet32_10percent.csv};
    \addlegendentry{ESPN-Finetune}
    
    \addplot[mark=x,red]
        table {tikz/rewind_resnet32_10percent.csv};
    \addlegendentry{ESPN-Rewind}
        
    \addplot[mark=square,green]
        table {tikz/lottery_resnet32_10percent.csv};
    \addlegendentry{LT}
        
    \addplot[mark=diamond,orange]
        table {tikz/snip_resnet32_10percent.csv};
    \addlegendentry{SNIP}
    
    \end{axis}
\end{tikzpicture}}}
        \vspace{0.5em}
        \subfloat[ResNet32 ($p=98\%$)]{\resizebox{0.415\columnwidth}{!}{\begin{tikzpicture}
    \begin{axis}[ymin=0,ymax=100,
        legend style={at={(0.75,0.8)},anchor=east},
        xlabel=Layer Index,
        axis background/.style={fill=blue!3},
        grid=both,
        grid style={line width=.1pt, draw=gray!10},
        major grid style={line width=.2pt,draw=gray!50},
        width=9cm,height=7cm]
    \addplot[mark=*,blue] 
        table {tikz/finetune_resnet32_2percent.csv};
    \addlegendentry{ESPN-Finetune}
    
    \addplot[mark=x,red]
        table {tikz/rewind_resnet32_2percent.csv};
    \addlegendentry{ESPN-Rewind}
        
    \addplot[mark=square,green]
        table {tikz/lottery_resnet32_2percent.csv};
    \addlegendentry{LT}
        
    \addplot[mark=diamond,orange]
        table {tikz/snip_resnet32_2percent.csv};
    \addlegendentry{SNIP}
    
    \end{axis}
\end{tikzpicture}}}
        \end{tabular}}
    \caption{\sl Weight distribution comparisons on ResNet32 with CIFAR10. Almost all pruning algorithms (including ESPN) tend to prune weights in the middle layers. Surprisingly though, ESPN also tends to prune initial layers, concentrating most non-zero weights in the final layers. We hypothesize that learning masks allows us to specifically learn sparser abstractions in earlier layers, analogous to learning sparse features in older classification techniques.}  
    \label{fig:weight_dist}
\end{figure}

To understand better why ESPN outperforms other existing pruning approach in extreme pruning ratio, we visualize sparsity ratios for each layer of the network. We analyse VGG19 and ResNet32 trained with CIFAR10 for two pruning ratios: $p=90\% \text{ and } 98\%$. 
Conventional pruning algorithms tend to remove fewer weights in earlier layers (to preserve fine features of the input) and prune more in the deeper layers with higher number of parameters. 
While VGG19 weight distributions from ESPN follows this trend, we observe that ESPN shows a different trend on ResNet32 compared to SNIP and Lottery Ticket hypothesis (\Figref{fig:weight_dist}). 
SNIP follows trend of conventional methods by pruning deeper layers aggressively while preserving weights in the beginning. LT's weight distribution is comparatively uniform across layers. 
Given that LT performs better than SNIP (refer \Secref{subsec:results}), we hypothesize that pruning deeper layers in ResNet32 aggressively may induce information bottlenecks, degrading the performance significantly. 
ESPN, counter-intuitively, prunes more in the earlier layers and the middle layers than in deeper layers. Considering that ESPN outperforms the SNIP and LT for most of the case, ESPN weight distribution counters the conventional pruning intuition and emphasizes the importance of careful rather than excessive pruning in deeper layers.

In case of VGG19, we observe opposite trend to ResNet32 on ESPN weight distribution. While ESPN-Finetune prunes more in the earlier layer and the middle layers than in deeper layers in ResNet32, ESPN aggressively prunes deeper layer compare to SNIP and LT. The presented weight distributions show that the pruning strategy using the same algorithm can differ based on the network architectures. 


\begin{figure}[htp!]
\centering
    \resizebox{.8\linewidth}{!}{
        \begin{tabular}{c c}
        \vspace{-0.5em}
        \subfloat[VGG19 ($p=90\%$)]{\resizebox{0.45\columnwidth}{!}{\begin{tikzpicture}
    \begin{axis}[ymin=0,ymax=100,
        legend style={at={(0.82,0.8)},anchor=east},
        xlabel=Layer Index,
        ylabel=Remaining Weights (\%),
        axis background/.style={fill=blue!3},
        grid=both,
        grid style={line width=.1pt, draw=gray!10},
        major grid style={line width=.2pt,draw=gray!50},
        width=9cm,height=7cm
        ]
    \addplot[mark=*,blue] 
        table {tikz/finetune_vgg19_10percent.csv};
    \addlegendentry{ESPN-Finetune}
    
    \addplot[mark=x,red]
        table {tikz/rewind_vgg19_10percent.csv};
    \addlegendentry{ESPN-Rewind}
        
    \addplot[mark=square,green]
        table {tikz/lottery_vgg19_10percent.csv};
    \addlegendentry{LT}
        
    \addplot[mark=diamond,orange]
        table {tikz/snip_vgg19_10percent.csv};
    \addlegendentry{SNIP}
    
    \end{axis}
\end{tikzpicture}}}
        \vspace{0.5em}
        \subfloat[VGG19 ($p=98\%$)]{\resizebox{0.415\columnwidth}{!}{\begin{tikzpicture}
    \begin{axis}[ymin=0,ymax=100,
        legend style={at={(0.75,0.8)},anchor=east},
        xlabel=Layer Index,
        axis background/.style={fill=blue!3},
        grid=both,
        grid style={line width=.1pt, draw=gray!10},
        major grid style={line width=.2pt,draw=gray!50},
        width=9cm,height=7cm]
    \addplot[mark=*,blue] 
        table {tikz/finetune_vgg19_2percent.csv};
    \addlegendentry{ESPN-Finetune}
    
    \addplot[mark=x,red]
        table {tikz/rewind_vgg19_2percent.csv};
    \addlegendentry{ESPN-Rewind}
        
    \addplot[mark=square,green]
        table {tikz/lottery_vgg19_2percent.csv};
    \addlegendentry{LT}
        
    \addplot[mark=diamond,orange]
        table {tikz/snip_vgg19_2percent.csv};
    \addlegendentry{SNIP}
    
    \end{axis}
\end{tikzpicture}}}
        \end{tabular}}
    \caption{\sl Weight distribution comparisons on VGG19 with CIFAR10. ESPN prunes aggressively in the first two layers and deeper layers. Considering ESPN outperforms SNIP, GraSP, and LT for the most of the case, good pruning strategies may be different depending on the architectures as ESPN's VGG19 weight distribution trend is opposite to the results from ResNet32.}  
    \label{fig:weight_dist2}
\end{figure}
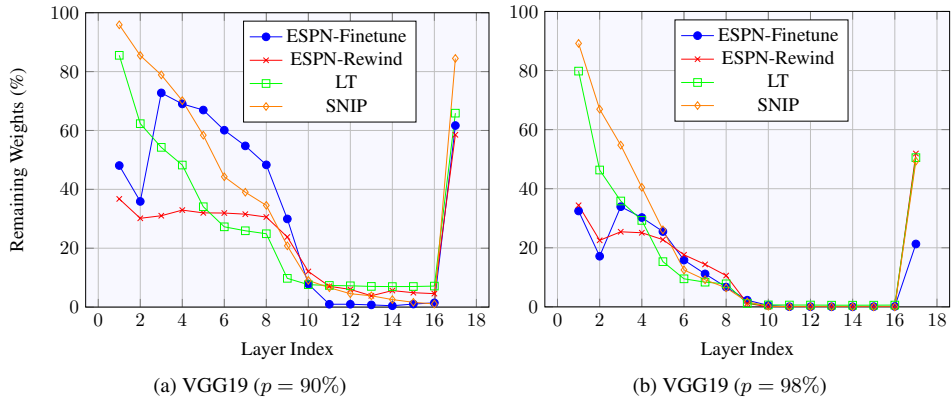


\subsection{Ablation Studies}
\label{subsection: ablation study}

\noindent\textbf{Role of weight updates, auxiliary parameters, and the L1 penalty.}
In the previous section, we have shown that our approach can prune the neural networks with various pruning ratios, specifically for extreme cases (> 99\%) with minor sparsity-accuracy tradeoff. We now analyse each of the components of ESPN through ablation studies. 


We consider four different scenarios while learning the auxiliary parameter $\rvc$: (1) updating both $\rvw$ and $\rvc$ with L1 penalty on $\rvc$ (original ESPN), (2) only updating $\rvc$ with L1 penalty on $\rvc$, (3) updating $\rvw$ and $\rvc$ without L1 penalty, and (4) only updating $\rvc$ without L1 penalty. Then we compare the test accuracy after the fine-tuning the model (Line 13).  We experiment on VGG19 with CIFAR10/100 datasets with pruning ratio \{70\%, 80\%, 90\%, 95\%, 95\%, 99\%, 99.5\%, 99.8\%, 99.9\%\} which includes common to extreme pruning ratio. We observe that freezing weights and optimizing for only $\rvc$ shows similar performance as ESPN, but is unstable for some pruning ratios. We also see improvement in performance when both $\rvw$ and $\rvc$ are updated. The results are shown in \Figref{appdxfig: Ablation Study ESPN}

\begin{figure}[htp!]
\centering
    \resizebox{.8\linewidth}{!}{
        \def\whist{0.37\linewidth}
        \def\whista{0.338\linewidth}
        \begin{tabular}{c c}
        \subfloat[VGG19 with CIFAR10]{\resizebox{\whist}{!}{\begin{tikzpicture}[]
    \begin{axis}[legend pos=south west,
        xlabel=Pruning ratio,
        ylabel=Test Accuracy,
        ymin=80,ymax=95,
        axis background/.style={fill=blue!3},
        grid=both,
        grid style={line width=.1pt, draw=gray!10},
        major grid style={line width=.2pt,draw=gray!50},
        width=10cm,height=7.5cm]
    \addplot[mark=*,blue] 
        plot coordinates {
            (70,93.67)
            (80,93.57)
            (90,93.7)
            (95,93.39)
            (99,92.26)
            (99.5,91.62)
            (99.8,89.23)
            (99.9,86.33)
        };
    \addlegendentry{Grad/L1}

    \addplot[mark=x,red] 
        plot coordinates {
            (70,92.58)
            (80,92.45)
            (90,92.66)
            (95,10.00)
            (99,92.59)
            (99.5,91.13)
            (99.8,89.32)
            (99.9,86.7)
        };
    \addlegendentry{No Grad/L1}

    \addplot[mark=square,green] 
        plot coordinates {
            (70,92.22)
            (80,92.22)
            (90,92.17)
            (95,92.19)
            (99,90.03)
            (99.5,82.02)
            (99.8,36.98)
            (99.9,10.0)
        };
    \addlegendentry{Grad/No L1}

    \addplot[mark=diamond,orange] 
        plot coordinates {
            (70,91.56)
            (80,91.22)
            (90,91.17)
            (95,90.93)
            (99,10.00)
            (99.5,10.00)
            (99.8,10.00)
            (99.9,10.00)
        };
    \addlegendentry{No Grad/No L1}
        
    \end{axis}
\end{tikzpicture}}}
        \vspace{0.7em}
        \subfloat[VGG19 with CIFAR100]{\resizebox{\whista}{!}{\begin{tikzpicture}
    \begin{axis}[legend pos=south west,
        xlabel=Pruning ratio,
        axis background/.style={fill=blue!3},
        grid=both,
        grid style={line width=.1pt, draw=gray!10},
        major grid style={line width=.2pt,draw=gray!50},
        width=10cm,height=7.5cm,
        ymin=55,ymax=77]

    \addplot[mark=*,blue] 
        plot coordinates {
            (70,73.95)
            (80,74.28)
            (90,73.18)
            (95,72.82)
            (99,68.76)
            (99.5,64.34)
            (99.8,56.71)
            (99.9,51.82)
        };
    \addlegendentry{Grad/L1}

    \addplot[mark=x,red] 
        plot coordinates {
            (70,72.78)
            (80,73.02)
            (90,73.32)
            (95,72.48)
            (99,69.16)
            (99.5,63.84)
            (99.8,55.11)
            (99.9,50.48)
        };
    \addlegendentry{No Grad/L1}

    \addplot[mark=square,green] 
        plot coordinates {
            (70,70.46)
            (80,70.42)
            (90,69.98)
            (95,69.16)
            (99,28.38)
            (99.5,2.42)
            (99.8,1.00)
            (99.9,1.00)
        };
    \addlegendentry{Grad/No L1}
[0.01, 0.01, 0.01, 0.01, 0.5741, 0.6412, 0.6561, 0.6568]
    \addplot[mark=diamond,orange] 
        plot coordinates {
            (70,65.68)
            (80,65.61)
            (90,64.12)
            (95,57.41)
            (99,1.00)
            (99.5,1.00)
            (99.8,1.00)
            (99.9,1.00)
        };
    \addlegendentry{No Grad/No L1}
        
    \end{axis}
\end{tikzpicture}}}
        \end{tabular}}
    \caption{Ablation study on four different setups learning the auxiliary parameters: ``Grad/No Grad" and ``L1/No L1'' correspond to weights updated or not and L1 penalty on auxiliary parameters or not, respectively in stage 1. We test from regular to extreme pruning ratio: \{70\%, 80\%, 90\%, 95\%, 99\%, 99.5\%, 99.8\%, 99.9\%\}.} 
    \label{appdxfig: Ablation Study ESPN}
\end{figure}
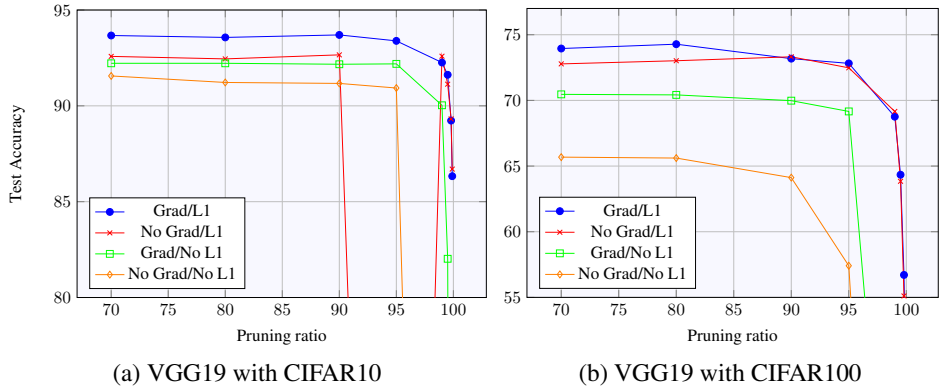

\begin{figure}[!h]
\centering
    \resizebox{.8\linewidth}{!}{
        \def\whist{0.34\columnwidth}
        \def\whista{0.30\columnwidth}
        \begin{tabular}{c c}
        \vspace{-0.5em}
        \subfloat[Learning Rate: 0.05]{\resizebox{\whist}{!}{\begin{tikzpicture}
    \begin{axis}[ymin=0,ymax=50000,
        legend pos=north east,
        xlabel=Epochs,
        ylabel=Non-Zero Elements,
        axis background/.style={fill=blue!3},
        grid=both,
        grid style={line width=.1pt, draw=gray!10},
        major grid style={line width=.2pt,draw=gray!50},]
    \addplot[mark=*,blue] 
        table {tikz/lr005_8e-5.csv};
    \addlegendentry{$\alpha=8\cdot 10^{-5}$}
    
    \addplot[mark=x,red]
        table {tikz/lr005_1e-4.csv};
    \addlegendentry{$\alpha=1\cdot10^{-4}$}
        
    \addplot[mark=square,green]
        table {tikz/lr005_2e-4.csv};
    \addlegendentry{$\alpha=2\cdot10^{-4}$}
        
    \addplot[mark=diamond,orange]
        table {tikz/lr005_3e-4.csv};
    \addlegendentry{$\alpha=3\cdot 10^{-4}$}
    
    \end{axis}
\end{tikzpicture}}}
        \vspace{0.5em}4
        \subfloat[Learning Rate: 0.1]{\resizebox{\whista}{!}{\begin{tikzpicture}
    \begin{axis}[ymin=0,ymax=50000,
        legend pos=north east,
        xlabel=Epochs,
        axis background/.style={fill=blue!3},
        grid=both,
        grid style={line width=.1pt, draw=gray!10},
        major grid style={line width=.2pt,draw=gray!50},]
    \addplot[mark=*,blue] 
        table {tikz/lr01_8e-5.csv};
    \addlegendentry{$\alpha=8\cdot10^{-5}$}
    
    \addplot[mark=x,red]
        table {tikz/lr01_1e-4.csv};
    \addlegendentry{$\alpha=1\cdot10^{-4}$}
        
    \addplot[mark=square,green]
        table {tikz/lr01_2e-4.csv};
    \addlegendentry{$\alpha=2\cdot10^{-4}$}
        
    \addplot[mark=diamond,orange]
        table {tikz/lr01_3e-4.csv};
    \addlegendentry{$\alpha=3\cdot10^{-4}$}
    
    \end{axis}
\end{tikzpicture}}}
        \vspace{0.5em}
        \subfloat[Learning Rate: 0.2]{\resizebox{\whista}{!}{\begin{tikzpicture}
    \begin{axis}[ymin=0,ymax=50000,
        legend pos=north east,
        xlabel=Epochs,
        axis background/.style={fill=blue!3},
        grid=both,
        grid style={line width=.1pt, draw=gray!10},
        major grid style={line width=.2pt,draw=gray!50},]
    \addplot[mark=*,blue] 
        table {tikz/lr02_8e-5.csv};
    \addlegendentry{$\alpha=8\cdot10^{-5}$}
    
    \addplot[mark=x,red]
        table {tikz/lr02_1e-4.csv};
    \addlegendentry{$\alpha=1\cdot10^{-4}$}
        
    \addplot[mark=square,green]
        table {tikz/lr02_2e-4.csv};
    \addlegendentry{$\alpha=2\cdot10^{-4}$}
        
    \addplot[mark=diamond,orange]
        table {tikz/lr02_3e-4.csv};
    \addlegendentry{$\alpha=3\cdot10^{-4}$}
    
    \end{axis}
\end{tikzpicture}}}
        \end{tabular}}
    \caption{Strength and rate of shrinkage comparison depending on lasso coefficient and learning rate with LeNet300.}  
    \label{fig: Lasso and Learning Rate}
\end{figure}
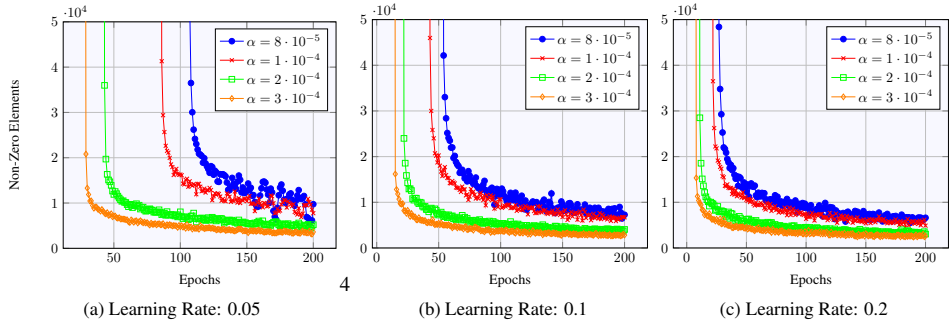

\subsection{Stability on Lasso Coefficient and Learning rate}
\label{appendix: additional experiments}

We check our algorithm's stability on lasso coefficient $\alpha$ and learning rate $\eta$ by observing the sparsity of $\rvc$ while learning the auxiliary parameters. We conduct experiments on tracking non-zero elements in $\rvc$ with various learning rate and lasso coefficient. We use pretrained LeNet300 (266K parameters) with MNIST dataset. Our experiment shows that both strength and rate of shrinkage on $\rvc$ are proportional to learning rate ($\eta$) and lasso coefficient $\alpha$. This shows that an extreme pruning ratio requires higher lasso coefficients to meet the condition sparsity of $\rvc$ less or equal to a targeted number of weights.

An interesting observation is that overall sparsity ratios post training are nearly independent of the learning rate, showing that the network is generally robust to reasonable hyperparameter choices. However, in the case of $\ell_1$ penalty, the choice of the coefficient needs to be high enough to ensure that we can achieve the required high sparsity.

\section{Discussion and Conclusions}

In this work, we provide a new algorithm, ESPN, which is a simple and scalable approach to prune a variety of neural network models. While ESPN achieves comparable (even improved) accuracy to SNIP, our algorithm is successfully able to compress the network to extremely high pruning levels ($>99\%$), up to the regime where the number of parameters are comparable to the input size. To the best of our knowledge, our approach is the first to achieve such high compression ratios for large networks such as ResNet32.

There still exist several open directions for further research. First, ESPN can perhaps be further expanded to a structured-pruning setup that enforces structural sparsity at level of either neurons, or convolutional filters, using group-lasso type constraints~\cite{yuan2007model}; this may provide the extra advantage of faster inference over unstructured pruning. Secondly, analysing ESPN's performance on large-scale language models such as GPT~\cite{radford2019language}/BERT~\cite{Devlin2019BERTPO} and other transformer models would be an important study to undertake. Finally, it would be interesting to carefully understand why our compressed models generalize so well, given that their high levels of compression.

\bibliographystyle{amsalpha}
\bibliography{neurips_2020}

\newcommand{\etalchar}[1]{$^{#1}$}
\providecommand{\bysame}{\leavevmode\hbox to3em{\hrulefill}\thinspace}
\providecommand{\MR}{\relax\ifhmode\unskip\space\fi MR }
\providecommand{\MRhref}[2]{%
  \href{http://www.ams.org/mathscinet-getitem?mr=#1}{#2}
}
\providecommand{\href}[2]{#2}
\begin{thebibliography}{RWC{\etalchar{+}}19}

\bibitem[AHS17]{anwar2017structured}
Sajid Anwar, Kyuyeon Hwang, and Wonyong Sung, \emph{Structured pruning of deep
  convolutional neural networks}, ACM Journal on Emerging Technologies in
  Computing Systems (JETC) \textbf{13} (2017), no.~3, 1--18.

\bibitem[BFG18]{belagiannis2018adversarial}
Vasileios Belagiannis, Azade Farshad, and Fabio Galasso, \emph{Adversarial
  network compression}, Euro. Conf. Comp. Vision, 2018, pp.~0--0.

\bibitem[BKML18]{bellec2017deep}
Guillaume Bellec, David Kappel, Wolfgang Maass, and Robert Legenstein,
  \emph{Deep rewiring: Training very sparse deep networks}, Proc. Int. Conf.
  Learning Representations (ICLR), 2018.

\bibitem[Cha89]{NIPS1988_133}
Yves Chauvin, \emph{A back-propagation algorithm with optimal use of hidden
  units}, Adv. Neural Inf. Proc. Sys. (NeurIPS) (D.~S. Touretzky, ed.),
  Morgan-Kaufmann, 1989, pp.~519--526.

\bibitem[CK14]{collins2014memory}
Maxwell~D Collins and Pushmeet Kohli, \emph{Memory bounded deep convolutional
  networks}, arXiv preprint arXiv:1412.1442 (2014).

\bibitem[CPI18]{carreira2018learning}
Miguel~A Carreira-Perpin{\'a}n and Yerlan Idelbayev,
  \emph{“learning-compression” algorithms for neural net pruning}, {IEEE}
  Conf. Comp. Vision and Pattern Recog, 2018, pp.~8532--8541.

\bibitem[CWT{\etalchar{+}}15]{chen2015hashednet}
Wenlin Chen, James~T. Wilson, Stephen Tyree, Kilian~Q. Weinberger, and Yixin
  Chen, \emph{Compressing neural networks with the hashing trick}, Proc. Int.
  Conf. Machine Learning (ICML), 2015.

\bibitem[CWZZ17]{cheng2017survey}
Yu~Cheng, Duo Wang, Pan Zhou, and Tao Zhang, \emph{A survey of model
  compression and acceleration for deep neural networks}, arXiv preprint
  arXiv:1710.09282 (2017).

\bibitem[DCLT19]{Devlin2019BERTPO}
Jacob Devlin, Ming-Wei Chang, Kenton Lee, and Kristina Toutanova, \emph{Bert:
  Pre-training of deep bidirectional transformers for language understanding},
  arXiv preprint, arXiV:1810.04805 (2019).

\bibitem[DSD{\etalchar{+}}13]{denil2013predicting}
Misha Denil, Babak Shakibi, Laurent Dinh, Marc'Aurelio Ranzato, and Nando
  De~Freitas, \emph{Predicting parameters in deep learning}, Adv. Neural Inf.
  Proc. Sys. (NeurIPS), 2013.

\bibitem[DZW18]{dai2018compressing}
Bin Dai, Chen Zhu, and David Wipf, \emph{Compressing neural networks using the
  variational information bottleneck}, Proc. Int. Conf. Machine Learning
  (ICML), 2018.

\bibitem[FC18]{Frankle2018TheLT}
Jonathan Frankle and Michael Carbin, \emph{The lottery ticket hypothesis:
  Finding sparse, trainable neural networks}, Proc. Int. Conf. Learning
  Representations (ICLR), 2018.

\bibitem[FDRC19]{Frankle2019TheLT}
Jonathan Frankle, Gintare~Karolina Dziugaite, Daniel~M. Roy, and Michael
  Carbin, \emph{Stabilizing the lottery ticket hypothesis}, arXiv preprint
  arXiv:1903.01611 (2019).

\bibitem[GRK17]{gray2017gpu}
Scott Gray, Alec Radford, and Diederik~P Kingma, \emph{Gpu kernels for
  block-sparse weights}, arXiv preprint arXiv:1711.09224 (2017).

\bibitem[GYC16]{guo2016dynamic}
Yiwen Guo, Anbang Yao, and Yurong Chen, \emph{Dynamic network surgery for
  efficient dnns}, 2016.

\bibitem[HCS{\etalchar{+}}17]{hubara2017quantized}
Itay Hubara, Matthieu Courbariaux, Daniel Soudry, Ran El-Yaniv, and Yoshua
  Bengio, \emph{Quantized neural networks: Training neural networks with low
  precision weights and activations}, J. Machine Learning Research \textbf{18}
  (2017), no.~1, 6869--6898.

\bibitem[HLL{\etalchar{+}}18]{he2018amc}
Yihui He, Ji~Lin, Zhijian Liu, Hanrui Wang, Li-Jia Li, and Song Han, \emph{Amc:
  Automl for model compression and acceleration on mobile devices}, Euro. Conf.
  Comp. Vision, 2018, pp.~784--800.

\bibitem[HMD15]{han2015deep}
Song Han, Huizi Mao, and William~J Dally, \emph{Deep compression: Compressing
  deep neural networks with pruning, trained quantization and huffman coding},
  Proc. Int. Conf. Learning Representations (ICLR), 2015.

\bibitem[HS93]{hassibi1993second}
Babak Hassibi and David~G Stork, \emph{Second order derivatives for network
  pruning: Optimal brain surgeon}, Adv. Neural Inf. Proc. Sys. (NeurIPS), 1993.

\bibitem[Kri09]{cifar10}
Alex Krizhevsky, \emph{Learning multiple layers of features from tiny images},
  Tech. report, 2009.

\bibitem[LAT19]{Lee2019SNIPSN}
Namhoon Lee, Thalaiyasingam Ajanthan, and Philip H.~S. Torr, \emph{Snip:
  Single-shot network pruning based on connection sensitivity}, Proc. Int.
  Conf. Learning Representations (ICLR), vol. abs/1810.02340, 2019.

\bibitem[LDS90]{lecun1990optimal}
Yann LeCun, John~S Denker, and Sara~A Solla, \emph{Optimal brain damage}, Adv.
  Neural Inf. Proc. Sys. (NeurIPS), 1990.

\bibitem[LLS{\etalchar{+}}17]{liu2017learning}
Zhuang Liu, Jianguo Li, Zhiqiang Shen, Gao Huang, Shoumeng Yan, and Changshui
  Zhang, \emph{Learning efficient convolutional networks through network
  slimming}, Proc. {IEEE} Intl. Conf. Comp. Vision, 2017, pp.~2736--2744.

\bibitem[LRLZ17]{lin2017runtime}
Ji~Lin, Yongming Rao, Jiwen Lu, and Jie Zhou, \emph{Runtime neural pruning},
  Adv. Neural Inf. Proc. Sys. (NeurIPS), 2017, pp.~2181--2191.

\bibitem[LSZ{\etalchar{+}}19]{liu2018rethinking}
Zhuang Liu, Mingjie Sun, Tinghui Zhou, Gao Huang, and Trevor Darrell,
  \emph{Rethinking the value of network pruning}, 2019.

\bibitem[LUW17]{louizos2017bayesian}
Christos Louizos, Karen Ullrich, and Max Welling, \emph{Bayesian compression
  for deep learning}, Adv. Neural Inf. Proc. Sys. (NeurIPS), 2017.

\bibitem[LWK18]{Louizos2018LearningSN}
Christos Louizos, Max Welling, and Diederik~P. Kingma, \emph{Learning sparse
  neural networks through l0 regularization}, 2018.

\bibitem[LWL17]{luo2017thinet}
Jian-Hao Luo, Jianxin Wu, and Weiyao Lin, \emph{Thinet: A filter level pruning
  method for deep neural network compression}, Proc. {IEEE} Intl. Conf. Comp.
  Vision, 2017, pp.~5058--5066.

\bibitem[MAV17]{molchanov2017variational}
Dmitry Molchanov, Arsenii Ashukha, and Dmitry Vetrov, \emph{Variational dropout
  sparsifies deep neural networks}, Proc. Int. Conf. Machine Learning (ICML),
  2017.

\bibitem[MMS{\etalchar{+}}18]{mocanu2018scalable}
Decebal~Constantin Mocanu, Elena Mocanu, Peter Stone, Phuong~H Nguyen,
  Madeleine Gibescu, and Antonio Liotta, \emph{Scalable training of artificial
  neural networks with adaptive sparse connectivity inspired by network
  science}, Nature communications \textbf{9} (2018), no.~1, 1--12.

\bibitem[MS89]{NIPS1988_119}
Michael~C Mozer and Paul Smolensky, \emph{Skeletonization: A technique for
  trimming the fat from a network via relevance assessment}, Adv. Neural Inf.
  Proc. Sys. (NeurIPS) (D.~S. Touretzky, ed.), Morgan-Kaufmann, 1989,
  pp.~107--115.

\bibitem[MTK{\etalchar{+}}17]{molchanov2016pruning}
Pavlo Molchanov, Stephen Tyree, Tero Karras, Timo Aila, and Jan Kautz,
  \emph{Pruning convolutional neural networks for resource efficient transfer
  learning}, Proc. Int. Conf. Learning Representations (ICLR), 2017.

\bibitem[MW19]{mostafa2019parameter}
Hesham Mostafa and Xin Wang, \emph{Parameter efficient training of deep
  convolutional neural networks by dynamic sparse reparameterization}, Proc.
  Int. Conf. Machine Learning (ICML), 2019.

\bibitem[OBSS20]{oktay2019model}
Deniz Oktay, Johannes Ball{\'e}, Saurabh Singh, and Abhinav Shrivastava,
  \emph{Model compression by entropy penalized reparameterization}, 2020.

\bibitem[PGM{\etalchar{+}}19]{pytorch}
Adam Paszke, Sam Gross, Francisco Massa, Adam Lerer, James Bradbury, Gregory
  Chanan, Trevor Killeen, Zeming Lin, Natalia Gimelshein, Luca Antiga, Alban
  Desmaison, Andreas Kopf, Edward Yang, Zachary DeVito, Martin Raison, Alykhan
  Tejani, Sasank Chilamkurthy, Benoit Steiner, Lu~Fang, Junjie Bai, and Soumith
  Chintala, \emph{Pytorch: An imperative style, high-performance deep learning
  library}, Adv. Neural Inf. Proc. Sys. (NeurIPS), 2019.

\bibitem[PPA18]{polino2018model}
Antonio Polino, Razvan Pascanu, and Dan Alistarh, \emph{Model compression via
  distillation and quantization}, Proc. Int. Conf. Learning Representations
  (ICLR), 2018.

\bibitem[RDS{\etalchar{+}}15]{ILSVRC15}
Olga Russakovsky, Jia Deng, Hao Su, Jonathan Krause, Sanjeev Satheesh, Sean Ma,
  Zhiheng Huang, Andrej Karpathy, Aditya Khosla, Michael Bernstein,
  Alexander~C. Berg, and Li~Fei-Fei, \emph{{ImageNet Large Scale Visual
  Recognition Challenge}}, Intl. J. Comp. Vision \textbf{115} (2015), no.~3,
  211--252.

\bibitem[RFC20]{renda2020comparing}
Alex Renda, Jonathan Frankle, and Michael Carbin, \emph{Comparing rewinding and
  fine-tuning in neural network pruning}, Proc. Int. Conf. Learning
  Representations (ICLR) (2020).

\bibitem[RWC{\etalchar{+}}19]{radford2019language}
Alec Radford, Jeff Wu, Rewon Child, David Luan, Dario Amodei, and Ilya
  Sutskever, \emph{Language models are unsupervised multitask learners}.

\bibitem[SB16]{sau2016deep}
Bharat~Bhusan Sau and Vineeth~N Balasubramanian, \emph{Deep model compression:
  Distilling knowledge from noisy teachers}, arXiv preprint arXiv:1610.09650
  (2016).

\bibitem[Set97]{setiono1997penalty}
Rudy Setiono, \emph{A penalty-function approach for pruning feedforward neural
  networks}, Neural computation \textbf{9} (1997), no.~1, 185--204.

\bibitem[SSM19]{savarese2019winning}
Pedro Savarese, Hugo Silva, and Michael Maire, \emph{Winning the lottery with
  continuous sparsification}, arXiv preprint arXiv:1912.04427 (2019).

\bibitem[SWR20]{sanh2020movement}
Victor Sanh, Thomas Wolf, and Alexander~M Rush, \emph{Movement pruning:
  Adaptive sparsity by fine-tuning}, arXiv preprint arXiv:2005.07683 (2020).

\bibitem[TKTH18]{theis2018faster}
Lucas Theis, Iryna Korshunova, Alykhan Tejani, and Ferenc Husz{\'a}r,
  \emph{Faster gaze prediction with dense networks and fisher pruning}, arXiv
  preprint arXiv:1801.05787 (2018).

\bibitem[TLFF18]{tartaglione2018learning}
Enzo Tartaglione, Skjalg Leps{\o}y, Attilio Fiandrotti, and Gianluca Francini,
  \emph{Learning sparse neural networks via sensitivity-driven regularization},
  Adv. Neural Inf. Proc. Sys. (NeurIPS), 2018, pp.~3878--3888.

\bibitem[TVDJ20]{Touvron2020FixingTT}
Hugo Touvron, Andrea Vedaldi, Matthijs Douze, and Herv{\'e} J{\'e}gou,
  \emph{Fixing the train-test resolution discrepancy: Fixefficientnet}, arXiv
  preprint, arXiv:2003.08237 (2020).

\bibitem[UMW17]{ullrich2017soft}
Karen Ullrich, Edward Meeds, and Max Welling, \emph{Soft weight-sharing for
  neural network compression}, Proc. Int. Conf. Learning Representations
  (ICLR), 2017.

\bibitem[VSF20]{verdenius2020pruning}
Stijn Verdenius, Maarten Stol, and Patrick Forr{\'e}, \emph{Pruning via
  iterative ranking of sensitivity statistics}, arXiv preprint arXiv:2006.00896
  (2020).

\bibitem[WZG20]{wang2020picking}
Chaoqi Wang, Guodong Zhang, and Roger Grosse, \emph{Picking winning tickets
  before training by preserving gradient flow}, Proc. Int. Conf. Learning
  Representations (ICLR) (2020).

\bibitem[XWR19]{xiao2019autoprune}
Xia Xiao, Zigeng Wang, and Sanguthevar Rajasekaran, \emph{Autoprune: Automatic
  network pruning by regularizing auxiliary parameters}, Adv. Neural Inf. Proc.
  Sys. (NeurIPS), 2019.

\bibitem[YL07]{yuan2007model}
Ming Yuan and Yi~Lin, \emph{Model selection and estimation in the gaussian
  graphical model}, Biometrika \textbf{94} (2007), no.~1, 19--35.

\bibitem[YMD{\etalchar{+}}15]{Yang2015DeepFC}
Zichao Yang, Marcin Moczulski, Misha Denil, Nando de~Freitas, Alexander~J.
  Smola, Le~Song, and Ziyu Wang, \emph{Deep fried convnets}, Proc. {IEEE} Intl.
  Conf. Comp. Vision, 2015.

\bibitem[ZG17]{zhu2017prune}
Michael Zhu and Suyog Gupta, \emph{To prune, or not to prune: exploring the
  efficacy of pruning for model compression}, arXiv preprint arXiv:1710.01878
  (2017).

\bibitem[ZLLY19]{zhou2019deconstructing}
Hattie Zhou, Janice Lan, Rosanne Liu, and Jason Yosinski, \emph{Deconstructing
  lottery tickets: Zeros, signs, and the supermask}, Adv. Neural Inf. Proc.
  Sys. (NeurIPS), 2019, pp.~3592--3602.

\end{thebibliography}


\end{document}